%% file: neurips_2025.tex
\documentclass{article}
\pdfoutput=1

\usepackage[preprint]{neurips_2025}

\usepackage[utf8]{inputenc} 
\usepackage[T1]{fontenc}    
\usepackage{hyperref}       
\usepackage{url}            
\usepackage{booktabs}       
\usepackage{amsfonts}       
\usepackage{nicefrac}       
\usepackage{microtype}      
\usepackage{xcolor}         

\definecolor{cvprblue}{rgb}{0.21,0.49,0.74}
\hypersetup{
    colorlinks=true,
    citecolor=cvprblue,
    linkcolor=blue,
    urlcolor=green
}
\usepackage{graphicx}
\usepackage{wrapfig}
\usepackage{float}
\usepackage{amssymb}
\usepackage{amsmath,amssymb,physics}
\usepackage{mathrsfs}

\usepackage{caption}
\usepackage{subcaption}
\usepackage{array}

\title{Learnable Burst-Encodable Time-of-Flight Imaging for High-Fidelity Long-Distance Depth Sensing}

\author{%
    Manchao Bao\textsuperscript{1}\\
	\texttt{manchaobao@smail.nju.edu.cn} \\
	\And
	Shengjiang Fang\textsuperscript{1}\\
	\texttt{shengjiangfang@smail.nju.edu.cn}
	\AND
	Tao Yue\textsuperscript{1}\\
	\texttt{yuetao@nju.edu.cn}
	\And
	Xuemei Hu\textsuperscript{1}\\
	\texttt{xuemeihu@nju.edu.cn}
}

\begin{document}

\maketitle

\vbox{%
	\vskip -0.26in
	\hsize\textwidth
	\linewidth\hsize
	\centering
	\normalsize
	\footnotemark[1]Nanjing University
	\vskip 0.3in
}

\input{sec/0_abstract}    
\input{sec/1_introduction}
\input{sec/2_related_work}

\input{sec/3_method}
\input{sec/4_experiment}

\input{sec/5_physical_exp}

\input{sec/6_conclusion}
{
    \small
    \bibliographystyle{unsrt}
    \bibliography{references}
}

\end{document}

%% file: sec/0_abstract.tex
\begin{abstract}
Long-distance depth imaging holds great promise for applications such as autonomous driving and robotics. Direct time-of-flight (dToF) imaging offers high-precision, long-distance depth sensing, yet demands ultra-short pulse light sources and high-resolution time-to-digital converters. In contrast, indirect time-of-flight (iToF) imaging often suffers from phase wrapping and low signal-to-noise ratio (SNR) as the sensing distance increases. In this paper, we introduce a novel ToF imaging paradigm, termed Burst-Encodable Time-of-Flight (BE-ToF), which facilitates high-fidelity, long-distance depth imaging. Specifically, the BE-ToF system emits light pulses in burst mode and estimates the phase delay of the reflected signal over the entire burst period, thereby effectively avoiding the phase wrapping inherent to conventional iToF systems. Moreover, to address the low SNR caused by light attenuation over increasing distances, we propose an end-to-end learnable framework that jointly optimizes the coding functions and the depth reconstruction network. A specialized double well function and first-order difference term are incorporated into the framework to ensure the hardware implementability of the coding functions. The proposed approach is rigorously validated through comprehensive simulations and real-world prototype experiments, demonstrating its effectiveness and practical applicability.

\end{abstract}

%% file: sec/1_introduction.tex
\section{Introduction}
Achieving high-precision depth imaging over long distances has remained a fundamental objective in fields such as computer vision, robotics, and autonomous systems. Time-of-flight (ToF) imaging~\cite{chen2023breaking, roriz2021automotive, zennaro2015performance}, as a key approach to depth imaging, can be further categorized into direct ToF (dToF) and indirect ToF (iToF) based on differences in working principles. Direct ToF imaging~\cite{delpy1988estimation} estimates depth by directly measuring the round-trip time of light, enabling high-precision and long-range sensing. Despite its advantages, this approach requires ultra-short pulsed light sources and high-resolution time-to-digital converters (TDCs), imposing stringent hardware demands that increase system complexity and cost, thereby limiting its practicality for widespread deployment. Indirect ToF systems~\cite{hansard2012time, li2014time, frank2009theoretical, meng2024itof}, in contrast, emit amplitude-modulated continuous wave (AMCW) signals and infer depth by analyzing the phase shift between the transmitted and received signals. Due to their relatively lower hardware complexity and cost, iToF systems offer a more practical and hardware-friendly solution. Nevertheless, existing iToF technologies face significant challenges in long-range imaging, primarily due to phase wrapping~\cite{horaud2016overview} and low signal-to-noise ratio (SNR) resulting from optical attenuation~\cite{li2022fisher}. To address the phase wrapping, dual-frequency modulation techniques~\cite{poujouly2002twofold, jongenelen2011analysis} have been proposed, albeit at the cost of increased computational complexity and stricter hardware synchronization requirements. Alternative approaches have sought to mitigate phase wrapping under single-frequency modulation by incorporating scene priors~\cite{hansard2012time, crabb2015fast}, however, these methods do not fundamentally resolve the intrinsic ambiguity introduced by periodic modulation.

In this paper, we propose a novel ToF imaging paradigm termed Burst-Encodable Time-of-Flight (BE-ToF). Our BE-ToF system employs a low-frequency burst mode to modulate and demodulate high-frequency light pulses, wherein the phase shift of the reflected signal spans the full range [0, 2$\pi$] within a single burst period. This facilitates high-fidelity, long-distance depth imaging using only single frequency modulation. Moreover, considering the significant variation in SNRs caused by the light-falloff, we propose an end-to-end learnable framework that jointly optimizes the coding functions and the depth reconstruction network, thereby ensuring high-precision depth estimation. In particular, we incorporate constraints based on double well function and first-order difference to ensure the hardware implementability of the learned coding functions. We evaluate our method on a synthetic dataset and compare it with conventional iToF approaches, including single-frequency and multi-frequency modulation techniques. Finally, we built a prototype system to prove the effectiveness of our method in real-world experiments.

In general, we make the following contributions:
\begin{itemize}

\item We present a novel Burst-Encodable Time-of-Flight imaging system that enables high-fidelity long-distance depth sensing using only a single modulation frequency, thereby fundamentally mitigating the issue of phase wrapping inherent in traditional iToF systems.

\item We propose an end-to-end learnable framework that jointly optimizes the coding functions and the depth reconstruction network to ensure high-precision depth estimation across varying distances.

\item We uniquely incorporate double well function and first-order difference as loss function to ensure the hardware implementability of the learnable coding functions.

\item We develop a prototype of our BE-ToF system and demonstrate its superior performance on both simulated datasets and real-world scenarios.

\end{itemize}

%% file: sec/2_related_work.tex
\section{Related Work}
\label{Related Work}

\paragraph{ToF imaging.}
ToF imaging has emerged as a prevalent and effective technique for depth acquisition. Direct ToF imaging enables long-range depth estimation by measuring the round-trip time of light pulses~\cite{mccarthy2025high}. However, achieving high-precision depth measurements with dToF imposes stringent requirements on the pulsed light source, typically necessitating pulse widths in the nanosecond or picosecond range~\cite{delpy1988estimation, diop2013improving, koyama2018220}. In addition, the system requires high-resolution TDCs with picosecond-level timing precision~\cite{perenzoni201664, xu2023near}. These demanding hardware specifications present substantial challenges for practical implementation and large-scale deployment. In contrast, indirect ToF imaging leverages cost-effective CMOS sensors to deliver high-resolution depth estimation. However, iToF often faces phase ambiguity caused by phase wrapping when performing long-range depth imaging. A viable approach to address this issue is the use of multi-frequency modulation~\cite{poujouly2002twofold, droeschel2010multi}, where low frequencies are employed to extend the maximum unambiguous range, while high frequencies ensure precise depth measurements. Hanto \textit{et al.}~\cite{hanto2023time} developed a novel ToF LiDAR range finder based on dual-modulation frequency switching to achieve depth imaging with an extended range. Su \textit{et al.}~\cite{su2018deep} propose an end-to-end time-of-flight imaging framework that enables high-quality depth reconstruction under multi-frequency modulation. However, multi-frequency modulation often results in increased hardware complexity and computational cost. In addition, various approaches such as amplitude correction~\cite{hansard2012time} and surface normal constraints~\cite{crabb2015fast} have been proposed to enable phase unwrapping from single-frequency measurements, which extremely rely on the scene prior. In this paper, we propose Burst-Encodable Time-of-Flight Imaging to fundamentally solve the phase wrapping issue in iToF, enabling high-fidelity, long-distance depth estimation.

\paragraph{End-to-end learning.}
End-to-end learning is a method aimed at jointly optimizing optical systems and reconstruction algorithms. Metzler \textit{et al.}~\cite{metzler2020deep,sun2020learning} obtain high dynamic range (HDR) images from a single shot by jointly optimizing the optical encoder and the electronic decoder. Nie \textit{et al.}~\cite{nie2018deeply} leverage an end-to-end network for hyperspectral reconstruction, enabling simultaneous learning of optimized camera spectral response functions and a mapping for spectral reconstruction. For dense 3D localization microscopy, Nehme \textit{et al.}~\cite{nehme2020deepstorm3d} proposed a deep STORM-based method to achieve end-to-end optimization of point spread function engineering and accurate 3D localization. To extend the depth of field (EDOF), Sitzmann \textit{et al.}~\cite{sitzmann2018end} proposed jointly optimizing the optical system and the reconstruction algorithm's parameters to achieve achromatic EDOF imaging. Guo \textit{et al.}~\cite{guo2024end} put forward an end-to-end network framework capable of jointly optimizing the encoding function and exposure time to improve the accuracy of fluorescence lifetime imaging. Besides, in iToF imaging, Chugunov \textit{et al.}~\cite{chugunov2021mask} proposed to jointly learn a microlens amplitude mask and an encoder-decoder network to reduce flying pixels in depth captures. Li \textit{et al.}~\cite{li2022fisher} put forward a Fisher-information guided framework for the joint optimization of the coding functions and the reconstruction network. Given the remarkable potential of end-to-end learning in elevating imaging performance, we propose an end-to-end learnable framework that jointly optimizes the coding functions and depth reconstruction network of our BE-ToF, ensuring high-quality depth performance across varying distances.

%% file: sec/3_method.tex
\section{Learnable Burst-Encodable Time-of-Flight Imaging}
\label{Learnable Burst-Encodable Time-of-Flight Imaging}

\begin{figure}[h]
  \centering
   \includegraphics[width=\linewidth]{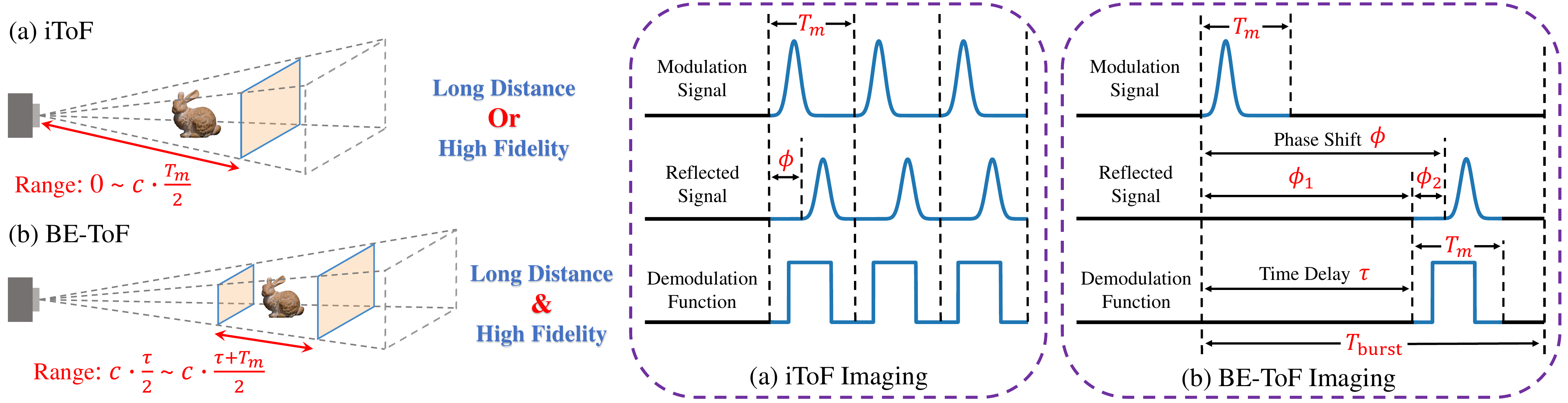}
   \caption{Comparison between iToF and BE-ToF. (a) Principle of iToF imaging, which suffers from a trade-off between sensing distance and precision; (b) Principle of BE-ToF imaging, enabling long-distance and high-fidelity depth sensing through modulation and demodulation in burst mode.}
   \label{fig:principle of BE-ToF imaging}
\end{figure}

In this section, we first introduce the working principle of our BE-ToF. The core idea of our BE-ToF is to modulate and demodulate high-frequency light pulses using a low-frequency burst mode. As illustrated in Fig.~\ref{fig:principle of BE-ToF imaging}(b), within a single long burst period $T_{burst}$, a short-cycle $T_m$ light pulse is emitted. When the reflected signal returns with a phase shift $\phi$, it can be demodulated by coding functions with controllable time delay. Specifically, the total phase shift $\phi$ can be decomposed into two components: $\phi_1$, which is primarily determined by the time delay $\tau$, and $\phi_2$, which can be recovered using demodulation techniques like 4-step phase shift or deep learning. In summary, the depth $d$ can be defined as Eq.~\ref{eq:depth}
\begin{equation}\label{eq:depth}
    d = \frac{c \: (\phi_1 + \phi_2) \: T_{\text{burst}}}{4\pi} 
    = \frac{c \tau}{2} + \mathcal{D}(\phi_2)
\end{equation}

where $c$ is the light speed and $\mathcal{D}(\phi_2)$ represents the demodulation process of $\phi_2$.

Thus, in our BE-ToF system, the maximum unambiguous range $d_{mur}$ is decided by $T_{burst}$, following $d_{mur} = \frac{c \cdot T_{burst}}{2}$, while depth accuracy $\epsilon_{d}$ is governed by $T_m$ as $\epsilon_{d} = \frac{c \cdot \epsilon_{\phi} \cdot T_m}{4\pi}$, where $\epsilon_{\phi}$ is the phase error during the demodulation process. Consequently, our BE-ToF system is capable of achieving both long-distance and high-precision depth imaging simultaneously. Furthermore, the depth sensing range in a single measurement spans from $\frac{c \cdot \tau}{2}$ to $\frac{c \cdot (\tau + T_m)}{2}$, which can be flexibly adjusted by tuning the time delay $\tau$.

\subsection{Differential BE-ToF Imaging Model}
\begin{figure}[h]
  \centering
   \includegraphics[width=\linewidth]{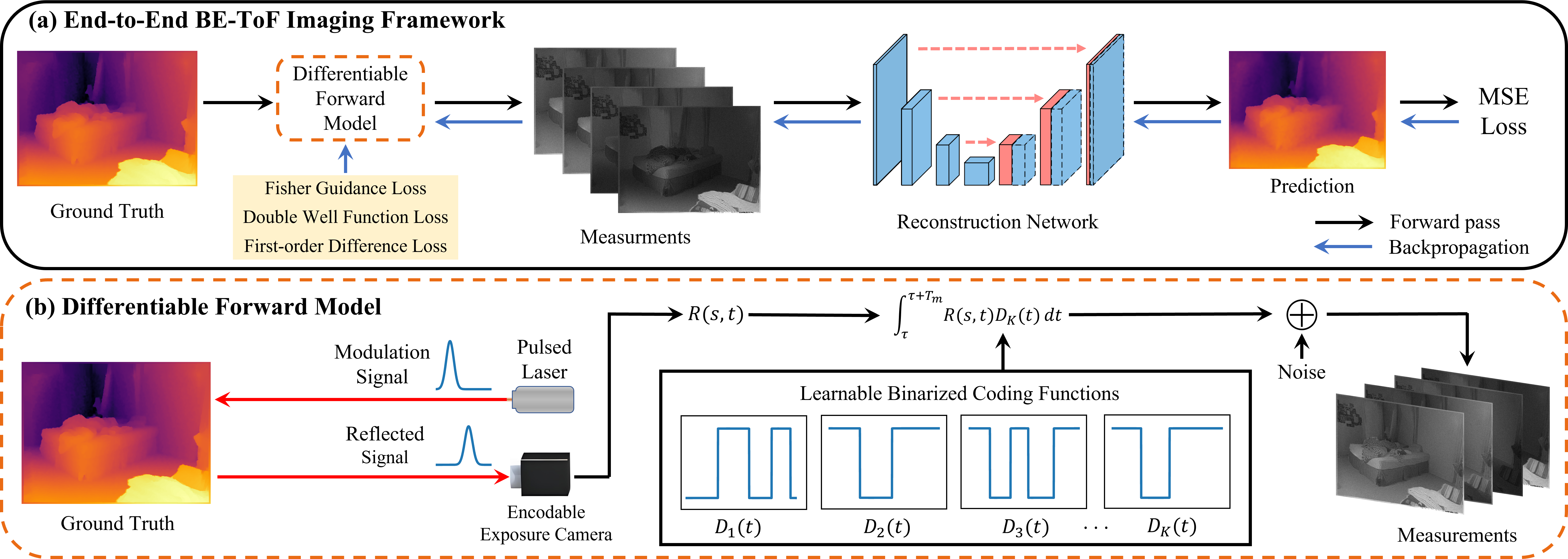}
   \caption{(a) End-to-end BE-ToF imaging framework for jointly optimizing the demodulation function and reconstruction network, (b) The differentiable physical model of the BE-ToF system.}
   \label{fig:overview}
\end{figure}

As described above, given the operating principle of BE-ToF, it can be readily implemented using a pulsed laser and an encodable exposure camera~\cite{agrawal2009coded}. Based on this, we first establish the differentiable physical model of our BE-ToF for end-to-end optimization. Assuming $M(t)$ is the modulation signal emitted by pulse laser, the reflected signal of scene point $s \in \mathbb{R}^{3}$ can be defined as Eq.~\ref{eq:reflected singal}
\begin{equation}\label{eq:reflected singal}
    R(s, t) = \rho_s M(t-2\frac{d(\mathbf{s})}{c}) + I_{amb}
\end{equation}

where $\rho_s$ is the inherent reflectance of the scene point $s$, $I_{amb}$ is the interference caused by ambient light, $d(s)$ denotes the depth value of point $s$. Furthermore, considering the attenuation of light intensity with distance during propagation, we incorporate the attenuation function into our model as Eq.~\ref{eq:singal fall off}
\begin{equation}\label{eq:singal fall off}
    R(s, t) = \mathcal{F}_{d(s)} \rho_s M(t-2\frac{d(s)}{c}) + I_{amb}
\end{equation}

where $\mathcal{F}_{d(s)}$ is the attenuation coefficient of the emitted light $M(t)$ at depth $d(s)$, which is typically inversely proportional to the square of the distance~\cite{liao2007light}. Finally, the whole BE-ToF imaging process can be formulated as Eq.~\ref{eq:measurements}
\begin{equation}\label{eq:measurements}
    I_i(s) = \int_{\tau}^{\tau+T_m} R(s, t)D_i(t) \, dt, \quad i \in 1,...,K
\end{equation}
where $I_i(s)$ is the measurement value of the camera, $D_i(t)$ denotes the coding functions and $K$ denotes the number of measurements. Taking into account the inherent noise of the sensor, the final measurement can be expressed as Eq.~\ref{eq:measurements with noise}
\begin{equation}\label{eq:measurements with noise}
    X_i(s) = I_i(s) + n_d + n_r, \;\;\; n_d \sim \mathcal{P}(\mathbb{E}(n_d)), \; n_r \sim \mathcal{N}(0, \sigma_{\mathrm{r}}^2)
\end{equation}
where $n_d$ is the dark noise following the poisson distribution with expectation $\mathbb{E}(n_d)$ and $n_r$ is the readout noise following gaussian distribution with standard deviation $\sigma_{\mathrm{r}}$. 

Considering that $X_i(s)$ contains three unknowns: $\rho_s$, $I_{amb}$, $\mathcal{F}_{d(s)}$. Therefore, at least $K \geq 3$ measurements are required to solve for the depth $d(s)$.

\subsection{Reconstruction Network}
\label{Reconstruction Network}

\begin{figure}[h]
  \centering
   \includegraphics[width=\linewidth]{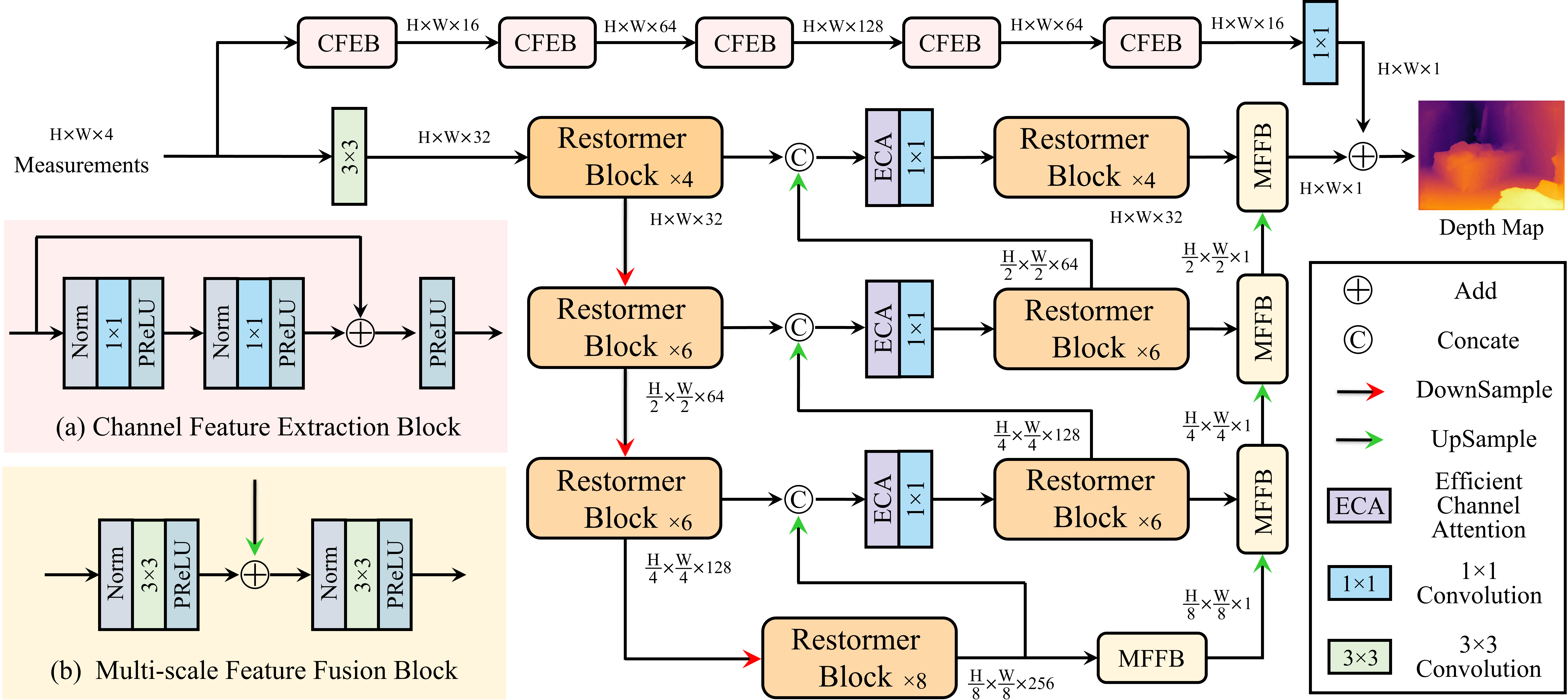}
   \caption{Architecture of the Restormer-based Spatial-Channel Fusion Network(RSCF-Net), (a) Channel Feature Extraction Block(CFEB), (b) Multi-scale Feature Fusion Block(MFFB).}
   \label{fig:Dual-branch Reconstruction Network}
\end{figure}

With the proposed differentiable forward model, we can simulate the $K$ measurements of the BE-ToF imaging process. To recover high-fidelity depth map from this set of measurements, we propose a Restormer-based Spatial-Channel Fusion Network(RSCF-Net). As shown in Fig.~\ref{fig:Dual-branch Reconstruction Network}, our network adopts Restormer~\cite{zamir2022restormer} as the backbone, featuring a four-level encoder-decoder structure in which each level comprises multiple Restormer blocks. In contrast to the conventional skip connections used in the original Restormer, we integrate an Efficient Channel Attention (ECA) module~\cite{wang2020eca} to enhance the fusion of features between encoder and decoder branches. Furthermore, recognizing the inherent differences between depth reconstruction and the image restoration tasks for which Restormer was originally designed, we augment our network with two additional components: the Channel Feature Extraction Block (CFEB) and the Multi-scale Feature Fusion Block (MFFB). The CFEB is composed of multiple residual-connected 1$\times$1 convolutional layers, designed to extract inter-channel relationships across multiple per-pixel measurements. On the other hand, the MFFB emphasizes spatial structure by performing preliminary depth estimation at each decoder level and progressively integrating features from multiple scales in a coarse-to-fine manner. The outputs of CFEB and MFFB are subsequently fused to produce the final high-fidelity depth map.

\subsection{Loss Function}
During the training process, we jointly optimize the coding functions and the reconstruction network. Given that our encodable exposure camera supports only binarized coding functions, we enforce hardware implementability by applying constraints based on a double well function and first-order difference. Additionally, Fisher information is incorporated into the loss to improve reconstruction quality, while mean squared error (MSE) is used as the objective to guide the final output. Here we give more details about these losses.

\begin{wrapfigure}{r}{0.35\linewidth}
   \centering
   \vspace{-1.2em}
   \includegraphics[width=0.98\linewidth]{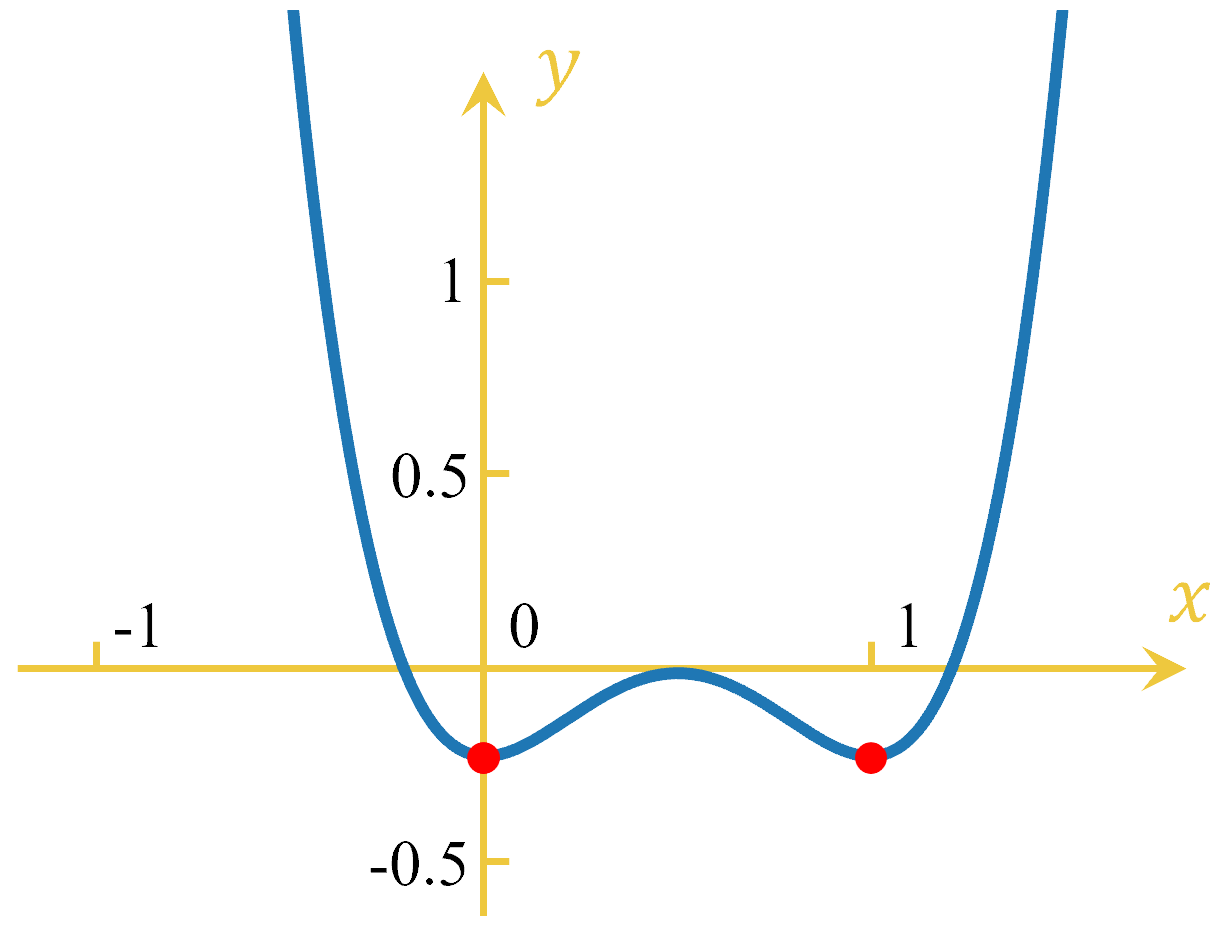}
   \caption{Demonstration of the double well function with two identical minima located at $x=0$ and $x=1$.}
   \label{fig:double_well_function}
   \vspace{-1em}
\end{wrapfigure}

\paragraph{Double Well Function Loss.}
To enable the optimization of binarized coding functions within the differentiable physical model. We introduce the double well function from quantum mechanics~\cite{jelic2012double}, which is formulated in Eq.~\ref{eq:double well function}
\begin{equation}\label{eq:double well function}
    f_{dw}(x) = 4(x-0.5)^4 - 2(x-0.5)^2
\end{equation}
As shown in Fig.~\ref{fig:double_well_function}, this function has two valleys at $x=0$ and $x=1$, thereby encouraging the coding functions to converge toward binary states during the optimization process. Therefore, our double well function loss can be defined as Eq.~\ref{eq:double well function loss}
\begin{equation}\label{eq:double well function loss}
    \mathcal{L}_{dw} = \sum_{i=1}^{K} \sum_{j=1}^{M} f_{dw}(D_i(t_j))
\end{equation}
where $M$ is the sampling points on each coding function.

\paragraph{First-order Difference Loss.}
Although the double well function effectively constrains the coding functions to a binary state, we observe that the learned functions often exhibit extremely narrow peaks, which pose challenges for practical hardware implementation. To mitigate this issue, we introduce a first-order difference loss, as defined in Eq.~\ref{eq:first-order difference loss}. By minimizing the first-order difference loss, narrow peaks can be effectively suppressed, thus ensuring feasibility for hardware implementation.
\begin{equation}\label{eq:first-order difference loss}
    \mathcal{L}_{1st} = \sum_{i=1}^{K} \sum_{j=1}^{M-1}\left| D_i(t_{j+1}) - D_i(t_j) \right|            
\end{equation}

\paragraph{Fisher Guidance Loss.}
The SNR is one of the key factors influencing the quality of ToF imaging. Inspired by~\cite{li2022fisher}, we introduce the fisher guidance loss to enhance the quality of our depth reconstruction, which can be summarized as Eq.~\ref{eq:fisher loss}
\begin{equation}\label{eq:fisher loss}
    \mathcal{L}_{fisher} = -\sum_{s} \sum_{i=1}^{K}  \left[ \frac{1}{2\sigma_{i}^{4}(s)} + \frac{1}{\sigma_{i}^{2}(s)} \right] \left[ \frac{\partial \mathbb{E}(I_i(s))}{\partial d} \right]^{2}
\end{equation}

where $\mathbb{E}(I_i(s))$ is the expectation of $I_i(s)$ and $\sigma_{i}(s) = \sqrt{\mathbb{E}(I_i(s)) + \mathbb{E}(n_d) + \sigma_{\mathrm{r}}^{2}}$.

\paragraph{Mean Squared Error Loss.}
We employ MSE as the fidelity loss to supervise the predicted depth map, as defined in Eq.~\ref{eq:MSE}
\begin{equation}\label{eq:MSE}
    \mathcal{L}_{MSE} = \sum_{s} \lVert d_{pre}(s) - d_{gt}(s) \rVert_{2}^{2}
\end{equation}

Finally, our complete loss can be summarized as Eq.~\ref{eq:complete loss}
\begin{equation}\label{eq:complete loss}
    \mathcal{L} = \mathcal{L}_{MSE} + \gamma_1 \mathcal{L}_{fisher} + \gamma_2 \mathcal{L}_{dw} + \gamma_3 \mathcal{L}_{1st}
\end{equation}

where $\gamma_1$, $\gamma_2$ and $\gamma_3$ are loss balance coefficients.

%% file: sec/4_experiment.tex
\begin{figure}[h] 
  \centering
   \includegraphics[width=\linewidth]{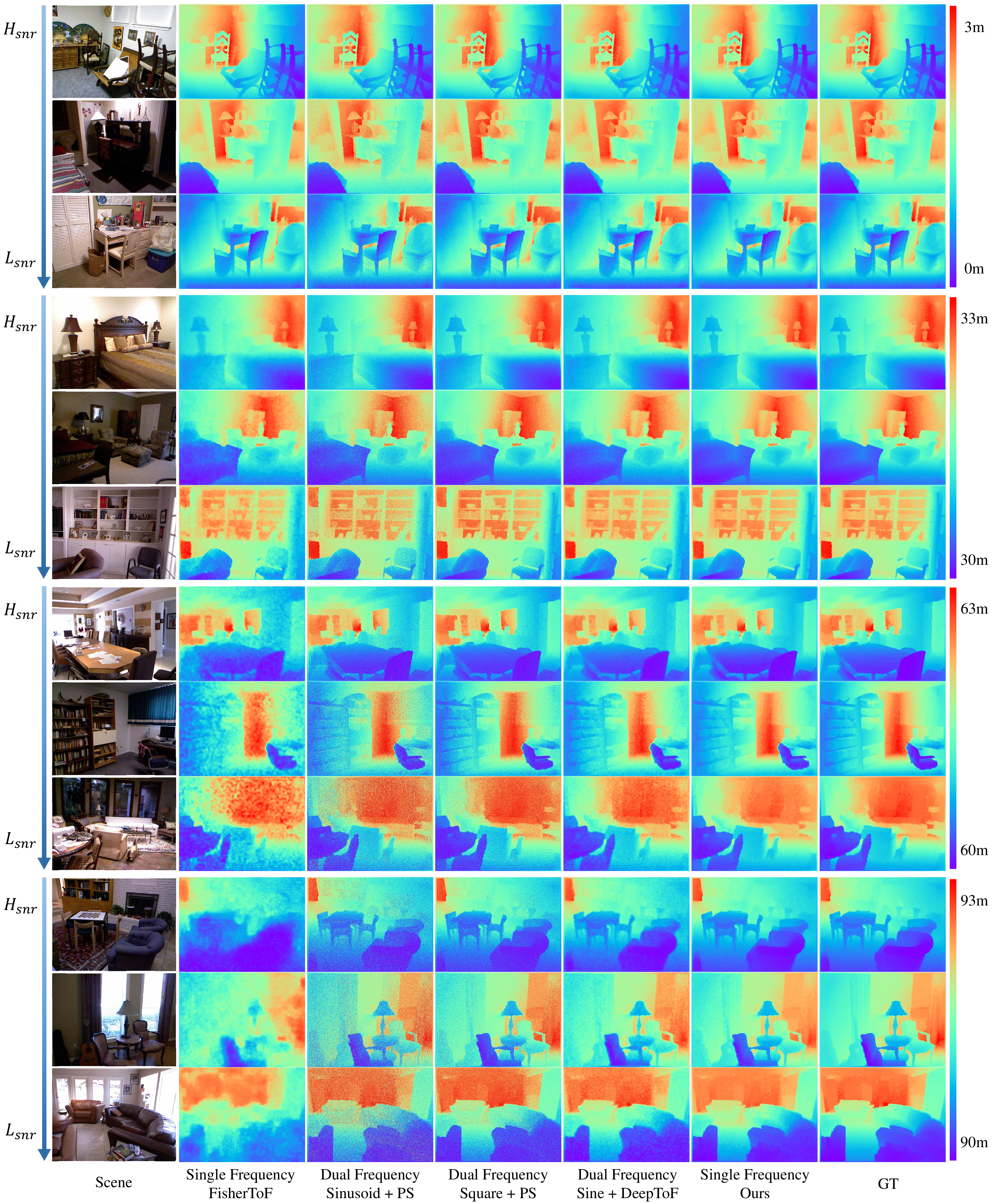}
   \caption{Overall comparisons with traditional iToF methods under various distances and SNRs, including FisherToF~\cite{li2022fisher} under single frequency modulation; Sine/Square + PS algorithm~\cite{hansard2012time} and Sine + DeepToF~\cite{su2018deep} under dual frequency modulation.}
    \vspace{-6mm}
   \label{fig:overall comparisons}
\end{figure}

\section{Synthetic Assessment}
\label{Synthetic Assessment}

\begin{table}
    \centering
    \caption{Quantitative comparison of overall performance, coding schemes, and reconstruction networks. All metrics are reported as MAE (mm).}
    \scriptsize
    \setlength{\tabcolsep}{5.5pt}
    \begin{tabular}{l *{12}{p{0.55cm}}}
        \toprule
          & \multicolumn{3}{c|}{0-3m} & \multicolumn{3}{c|}{30-33m} & \multicolumn{3}{c|}{60-63m} & \multicolumn{3}{c}{90-93m} \\
        \cmidrule(lr){2-4} \cmidrule(lr){5-7} \cmidrule(lr){8-10} \cmidrule(lr){11-13}
          & \multicolumn{1}{c}{$H_{\text{snr}}$} & \multicolumn{1}{c}{$M_{\text{snr}}$} & \multicolumn{1}{c|}{$L_{\text{snr}}$} & \multicolumn{1}{c}{$H_{\text{snr}}$} & \multicolumn{1}{c}{$M_{\text{snr}}$} & \multicolumn{1}{c|}{$L_{\text{snr}}$} & \multicolumn{1}{c}{$H_{\text{snr}}$} & \multicolumn{1}{c}{$M_{\text{snr}}$} & \multicolumn{1}{c|}{$L_{\text{snr}}$} & \multicolumn{1}{c}{$H_{\text{snr}}$} & \multicolumn{1}{c}{$M_{\text{snr}}$} & \multicolumn{1}{c}{$L_{\text{snr}}$} \\
        \midrule
        \multicolumn{13}{l}{\textbf{(a) Overall Performance}} \\ 
        \midrule
        Sine+PS~\cite{hansard2012time} & \multicolumn{1}{|c}{43.26} & \multicolumn{1}{c}{58.08} & \multicolumn{1}{c|}{78.09} & \multicolumn{1}{c}{56.96} & \multicolumn{1}{c}{77.89} & \multicolumn{1}{c|}{107.25} & \multicolumn{1}{c}{79.20} & \multicolumn{1}{c}{111.49} & \multicolumn{1}{c|}{158.40} & \multicolumn{1}{c}{117.21} & \multicolumn{1}{c}{170.76} & \multicolumn{1}{c}{244.29}\\
        Square+PS~\cite{hansard2012time} & \multicolumn{1}{|c}{33.21} & \multicolumn{1}{c}{40.64} & \multicolumn{1}{c|}{51.29} & \multicolumn{1}{c}{40.06} & \multicolumn{1}{c}{51.19} & \multicolumn{1}{c|}{66.90} & \multicolumn{1}{c}{51.90} & \multicolumn{1}{c}{69.12} & \multicolumn{1}{c|}{93.62} & \multicolumn{1}{c}{72.16} & \multicolumn{1}{c}{100.12} & \multicolumn{1}{c}{140.98}\\
        DeepToF~\cite{su2018deep} & \multicolumn{1}{|c}{17.76} & \multicolumn{1}{c}{19.19} & \multicolumn{1}{c|}{29.51} & \multicolumn{1}{c}{26.54} & \multicolumn{1}{c}{29.25} & \multicolumn{1}{c|}{37.49} & \multicolumn{1}{c}{31.50} & \multicolumn{1}{c}{35.02} & \multicolumn{1}{c|}{45.54} & \multicolumn{1}{c}{42.64} & \multicolumn{1}{c}{45.12} & \multicolumn{1}{c}{56.97}\\
        FisherToF~\cite{li2022fisher} & \multicolumn{1}{|c}{7.19} & \multicolumn{1}{c}{10.46} & \multicolumn{1}{c|}{16.91} & \multicolumn{1}{c}{20.80} & \multicolumn{1}{c}{24.54} & \multicolumn{1}{c|}{31.61} & \multicolumn{1}{c}{34.58} & \multicolumn{1}{c}{42.43} & \multicolumn{1}{c|}{54.73} & \multicolumn{1}{c}{75.19} & \multicolumn{1}{c}{77.68} & \multicolumn{1}{c}{138.11}\\
        \midrule
        \multicolumn{13}{l}{\textbf{(b) Coding Scheme}} \\ 
        \midrule
        Square & \multicolumn{1}{|c}{12.66} & \multicolumn{1}{c}{15.18} & \multicolumn{1}{c|}{21.35} & \multicolumn{1}{c}{16.82} & \multicolumn{1}{c}{22.10} & \multicolumn{1}{c|}{29.05} & \multicolumn{1}{c}{20.85} & \multicolumn{1}{c}{24.54} & \multicolumn{1}{c|}{32.77} & \multicolumn{1}{c}{25.51} & \multicolumn{1}{c}{30.08} & \multicolumn{1}{c}{44.47}\\
        \midrule
        \multicolumn{13}{l}{\textbf{(c) Reconstruction Network}} \\ 
        \midrule
        DeepToF~\cite{su2018deep} & \multicolumn{1}{|c}{14.76} & \multicolumn{1}{c}{16.20} & \multicolumn{1}{c|}{20.32} & \multicolumn{1}{c}{18.25} & \multicolumn{1}{c}{23.30} & \multicolumn{1}{c|}{34.12} & \multicolumn{1}{c}{24.95} & \multicolumn{1}{c}{31.50} & \multicolumn{1}{c|}{37.90} & \multicolumn{1}{c}{28.12} & \multicolumn{1}{c}{38.17} & \multicolumn{1}{c}{48.55}\\
        MaskToF~\cite{chugunov2021mask} & \multicolumn{1}{|c}{11.73} & \multicolumn{1}{c}{12.89} & \multicolumn{1}{c|}{17.41} & \multicolumn{1}{c}{14.72} & \multicolumn{1}{c}{19.44} & \multicolumn{1}{c|}{26.73} & \multicolumn{1}{c}{15.94} & \multicolumn{1}{c}{21.38} & \multicolumn{1}{c|}{33.55} & \multicolumn{1}{c}{25.71} & \multicolumn{1}{c}{27.86} & \multicolumn{1}{c}{38.60}\\
        FisherToF~\cite{li2022fisher} & \multicolumn{1}{|c}{6.94} & \multicolumn{1}{c}{8.10} & \multicolumn{1}{c|}{16.26} & \multicolumn{1}{c}{10.22} & \multicolumn{1}{c}{15.71} & \multicolumn{1}{c|}{21.68} & \multicolumn{1}{c}{14.62} & \multicolumn{1}{c}{18.31} & \multicolumn{1}{c|}{29.73} & \multicolumn{1}{c}{22.60} & \multicolumn{1}{c}{24.89} & \multicolumn{1}{c}{31.83}\\
        \midrule
        \textbf{Ours} & \multicolumn{1}{|c}{\textbf{5.90}} & \multicolumn{1}{c}{\textbf{6.95}} & \multicolumn{1}{c|}{\textbf{12.71}} & \multicolumn{1}{c}{\textbf{8.03}} & \multicolumn{1}{c}{\textbf{12.25}} & \multicolumn{1}{c|}{\textbf{18.29}} & \multicolumn{1}{c}{\textbf{11.93}} & \multicolumn{1}{c}{\textbf{16.60}} & \multicolumn{1}{c|}{\textbf{26.08}} & \multicolumn{1}{c}{\textbf{18.96}} & \multicolumn{1}{c}{\textbf{21.99}} & \multicolumn{1}{c}{\textbf{29.58}}\\
        \bottomrule
    \end{tabular}
    \label{tab:overall comparisons}
\end{table}

\subsection{Implementation Details}
\label{Implementation Details}
\paragraph{Dataset.} 
We use the NYU-V2 dataset~\cite{silberman2012indoor} to train and test our end-to-end framework. The NYU-V2 dataset is a high-quality RGB-D dataset captured by Kinect with a resolution of 640$\times$480. It contains a total of 1449 pairs of precisely aligned RGB and depth images collected from 464 indoor scenes, which enables its extensive application in academic research. For each RGB-D pair, we initially employ intrinsic image decomposition~\cite{jeon2014intrinsic} to break down the RGB image into albedo map and shading map. Subsequently, we designate the R-channel of the albedo map as the albedo component $\rho_s$ mentioned in Eq.~\ref{eq:reflected singal}, and the average of the three RGB channels of the albedo map is taken as the ambient component $I_{amb}$ in the same equation. We divide the dataset in detail, using 1000 pairs of data as the training set and the remaining 449 pairs as the test set~\cite{kim2021deformable,he2021towards}.

\paragraph{Incremental Training Method.} 
In our BE-ToF system, the SNR varies not only with distance but also significantly under the same distance due to ambient light $I_{amb}$. Therefore, we introduce an incremental training strategy~\cite{bengio2009curriculum} to ensure robust depth estimation of our network under varying SNR levels. Specifically, for each distance, we define three distinct SNR scenarios arranged from high to low. The network is trained with input data of varying SNRs, progressively transitioning from high to low every 10 epochs. When data of all SNRs are traversed, samples with random SNR are generated and fed to the network for the convergence of the network.

\paragraph{Training Parameters.} 
We choose $K$ in Eq.~\ref{eq:measurements} as 4 and $M$ in Eq.~\ref{eq:double well function loss} as 1000. The number of restormer blocks in the network is set to $[4, 6, 6, 8]$. We train the network for 200 epochs using the ADAM optimizer~\cite{kingma2014adam} with a batch size of 20. The learning rate is initialized at 0.01 and decays by a factor of 0.7 every 10 epochs. The loss balance coefficients $\gamma_1$ and $\gamma_2$ are empirically set to 5e-4 and 5e-2 initially, and are updated to 5e-5 and 1 after 40 epochs. $\gamma_3$ is always set to 5. Xavier initialization is used for the learnable coding functions. All experiments are conducted on the PyTorch platform~\cite{paszke2017automatic}, using an NVIDIA GeForce RTX 4090 GPU.

\subsection{Comparison with the State-of-the-art Methods}
To demonstrate the superiority of our method, we conduct a detailed comparison with traditional iToF approaches, including single frequency modulation and dual frequency modulation. The scenarios encompass multiple distance ranges (0-3\,m, 30-33\,m, 60-63\,m, and 90-93\,m) combined with varying SNRs, specifically high ($H_{snr}$ = 5.23\,dB), medium($M_{snr}$ = 3.68\,dB) and low ($L_{snr}$ = 2.22\,dB). As shown in Fig.~\ref{fig:overall comparisons}, we first compare our method with FisherToF~\cite{li2022fisher} under single frequency modulation. While FisherToF achieves precise depth reconstruction at close range, it still suffers from the rapid decline in imaging quality over distance. We then compare our method with a variety of dual frequency modulation approaches, including sinusoid and square coding functions with Phase Shift (PS) algorithm~\cite{hansard2012time} and the learning-based DeepToF~\cite{su2018deep} method. Our method achieves the best performance across various distances and SNRs, using only single frequency modulation. We present a detailed comparison in Tab.~\ref{tab:overall comparisons} (a), with Mean Absolute Error (MAE) as the evaluation criterion.

\begin{figure}[t] 
  \centering
   \includegraphics[width=\linewidth]{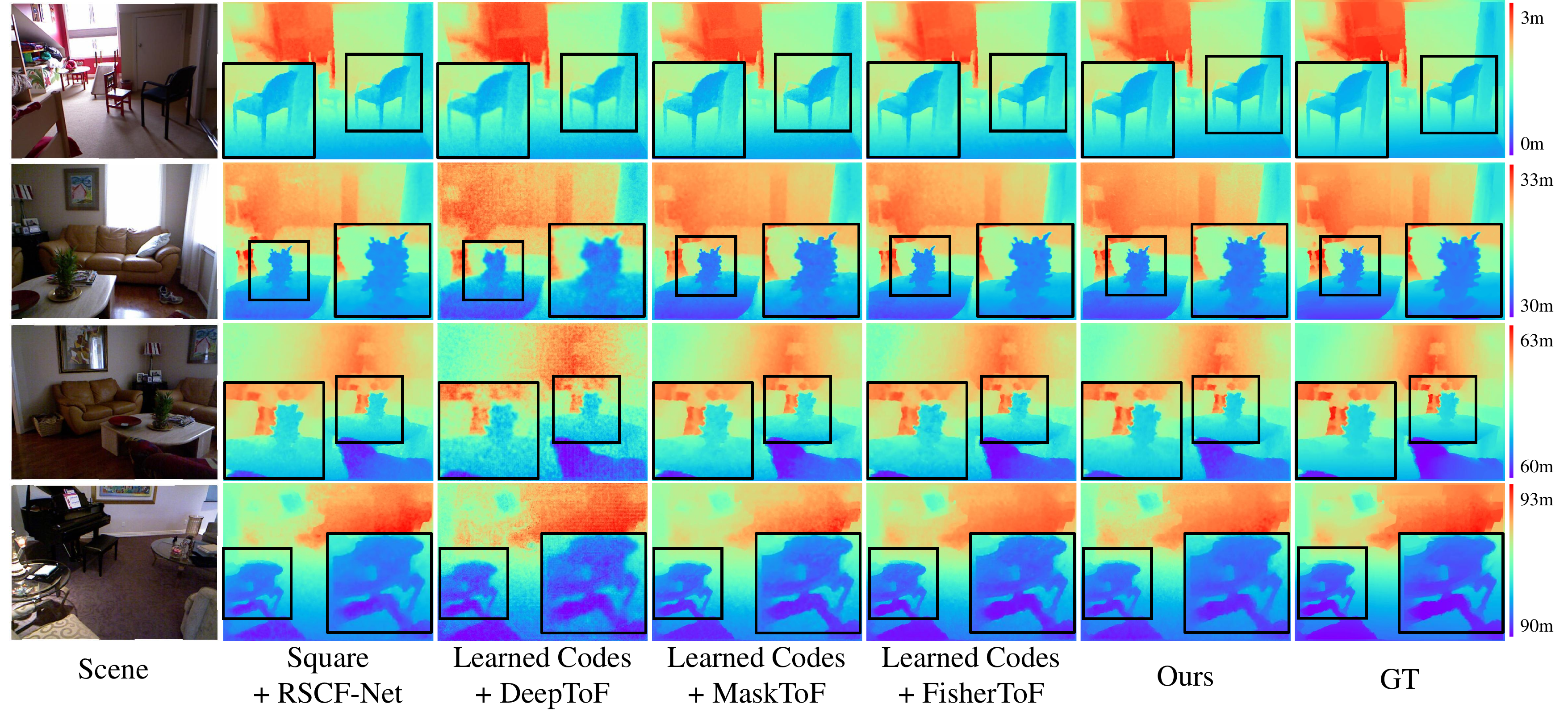}
   \caption{Comparisons with different coding scheme and reconstruction networks.}
   \label{fig:comparison with coding scheme and network}
\end{figure}

We further substantiate the superiority of our method through an analysis of the learnable coding function and the proposed RSCF-Net. As for the coding function, considering the practical hardware implementability, we compare our learning coding functions with the square encoding functions with the same RSCF-Net. The quantitative results presented in Tab.~\ref{tab:overall comparisons} (b) prove that our learning encoding functions provides superior depth reconstruction and enhanced robustness to noise. Additionally, we perform a thorough comparison of our RSCF-Net with existing depth reconstruction networks with the same learned coding functions, including DeepToF~\cite{su2018deep}, MaskToF~\cite{chugunov2021mask} and FisherToF~\cite{li2022fisher}. The quantitative results in Tab.~\ref{tab:overall comparisons} (c) confirm the effectiveness of our network. Fig.~\ref{fig:comparison with coding scheme and network} presents the visual results of different methods across four distances under low SNR conditions, intuitively demonstrating the advantages of our approach.

\subsection{Ablation Study}

\begin{figure}[htbp]
  \centering
  \begin{minipage}{0.6\textwidth}
    \centering
    \includegraphics[width=\linewidth]{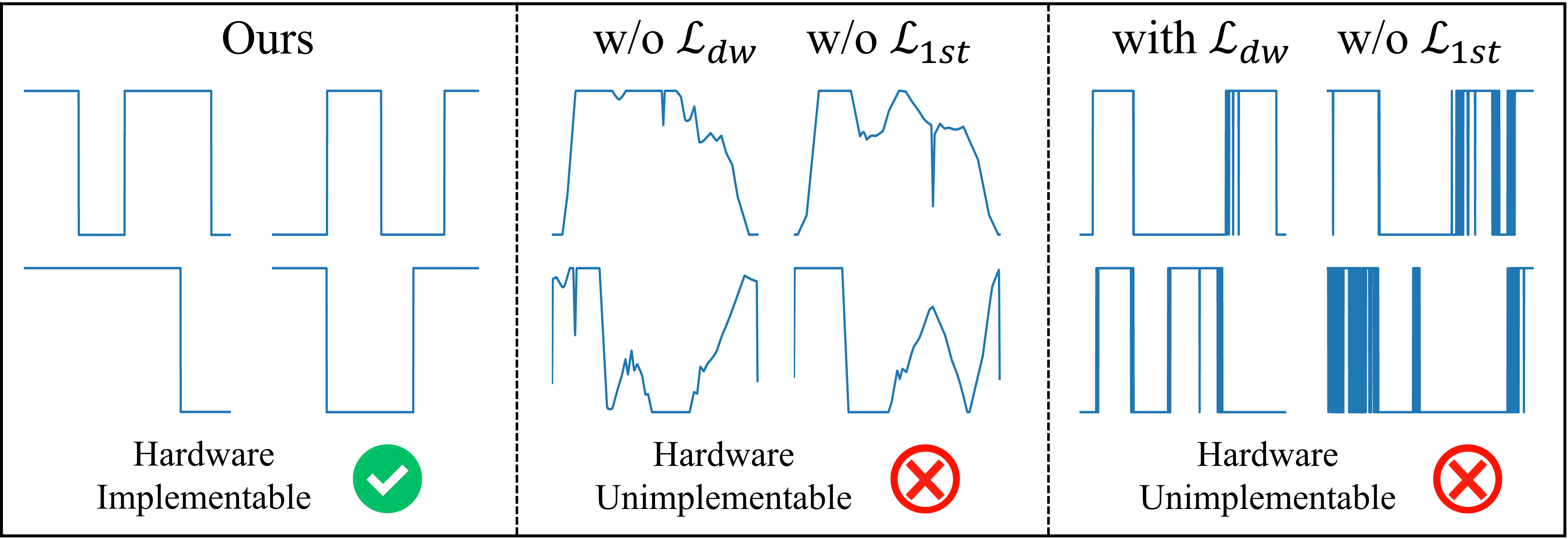}
    \caption{Visual ablations on $\mathcal{L}_{dw}$ and $\mathcal{L}_{1st}$.}
    \label{fig:ablation}
  \end{minipage}
  \hfill
  \begin{minipage}{0.37\textwidth}
    \centering
    \scriptsize
    \captionof{table}{Quantitative ablations with MAE(mm) as the evaluation metric.}
    \label{ablation}
    \setlength{\tabcolsep}{5pt}
    \begin{tabular}{@{}lcccc@{}}
      \toprule
      Distance(m) & 0-3 & 30-33 & 60-63 & 90-93 \\
      \midrule
      w/o $\mathcal{L}_{fisher}$   & 15.36 & 19.41 & 27.02 & 37.45 \\
      w/o CFEB                     & 58.69 & 73.64 & 78.56 & 86.94 \\
      w/o MFFB                     & 9.8 & 15.77 & 23.36 & 28.40 \\
      w/o ECA                      & 10.22 & 13.01 & 21.41 & 26.06 \\
      \textbf{Ours}                & \textbf{8.52} & \textbf{12.86} & \textbf{18.20} & \textbf{23.51} \\
      \bottomrule
    \end{tabular}
    \label{tab:ablation}
  \end{minipage}
\end{figure}

We first conduct ablation experiments on the proposed double well function loss and first-order difference loss. Since the learned coding functions are primarily used to control the camera's exposure, the absence of $\mathcal{L}_{dw}$ and $\mathcal{L}_{1st}$ results in coding functions that are entirely impractical to implement in hardware, which can be clearly illustrated in Fig.~\ref{fig:ablation}. In the next, we present a quantitative analysis in Tab.~\ref{tab:ablation} to evaluate the impact of the fisher guidance loss and different network blocks. The values in Tab.~\ref{tab:ablation} represent the average MAE measured under different SNRs at the same distance. The experimental results demonstrate the effectiveness of the introduced Fisher loss in guiding the network to learn an optimal coding functions. The ablation studies on different network blocks further validate the significant improvement in reconstruction quality brought by the proposed CFEB and MFFB.

%% file: sec/5_physical_exp.tex
\section{Physical Experiment Results}
\begin{figure}[h] 
  \centering
   \includegraphics[width=\linewidth]{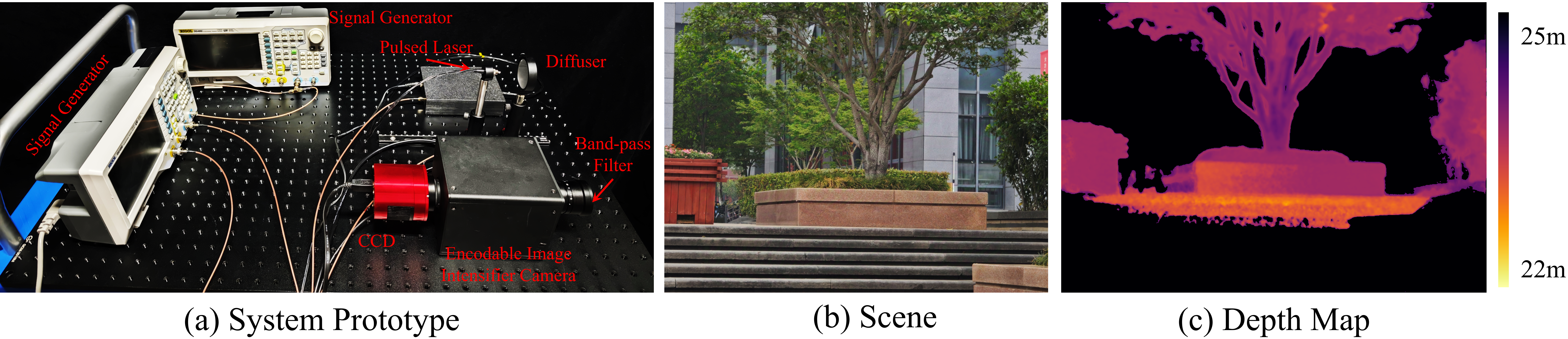}
   \caption{(a) System prototype of the BE-ToF, (b) Outdoor Scene, (c) Depth map.}
   \label{fig:real_world}
\end{figure}

\paragraph{Hardware Prototype.}
We develop a prototype system to validate the effectiveness of the proposed BE-ToF method in real-world scenarios. As illustrated in Fig.~\ref{fig:real_world}(a), the proposed system primarily comprises a cost-efficient fiber-based pulsed laser and an encodable image intensifier camera. The laser operates at a wavelength of 905\,nm, with a pulse width of 20\,ns and an average output power of 100\,mW. It features a numerical aperture (NA) of 0.22, enabling wide-area illumination suitable for array-based depth imaging. A diffuser is placed at the laser output to ensure uniform illumination across the scene. The image intensifier camera is composed of a primary lens (75\,mm, KOWA), image intensifier tubes, and a CCD sensor (ASI294MC), allowing for the detection of weak optical signals. To suppress ambient light interference, a 905\,nm band-pass filter (FWHM $\pm$ 10\,nm) is mounted in front of the lens. Both the laser and the image intensifier camera are driven in a 200\,kHz burst mode using a function generator (DG4202, Rigol), with synchronization handled by a second generator. The demodulation function is configured with a 50\,ns period, with the time delay flexibly adjustable via the function generator.

\paragraph{Experimental Results.}
In the real-world experiment, we selected a highly challenging scene located approximately 23 meters away, featuring complex vegetation, surfaces with varying reflectivity, and significant ambient light interference. The resulting depth map is presented in Fig.~\ref{fig:real_world}(c), where the depths of objects such as steps, flower beds, and trees are clearly distinguishable. Notably, our system is also capable of reconstructing the continuous depth profile of marble surfaces. It is important to note that the BE-ToF system captures depth within a limited range in a single measurement. However, this constraint can be advantageous, as it effectively filters out background clutter and enables the system to concentrate on the depth region of interest. As illustrated in Fig.~\ref{fig:real_world}(c), both foreground and background elements are successfully suppressed, allowing for a clear focus on the target range.

%% file: sec/6_conclusion.tex
\section{Conclusion, Limitations, and Broader Impact}
\label{Conclusion, Limitations, and Broader Impact}
In conclusion, we propose a novel ToF imaging paradigm, termed BE-ToF. The BE-ToF system enables long-distance high-fidelity depth imaging by modulating and demodulating pulsed signals in burst mode using only single-frequency modulation. Additionally, we introduce a learnable end-to-end framework that jointly optimizes binarized coding functions and the reconstruction network to effectively handle varying SNRs across different distances, achieving state-of-the-art performance.

\paragraph{Limitations.} 
Despite achieving both long-distance and high-fidelity depth imaging, our BE-ToF system is subject to limitations in its imaging range. As shown in Fig.~\ref{fig:principle of BE-ToF imaging}, the operational range is confined between $\frac{c \cdot \tau}{2}$ and $\frac{c \cdot (\tau + T_m)}{2}$, with higher precision resulting in a narrower imaging range.

\paragraph{Broader Impact.}
The proposed BE-ToF system demonstrates strong potential for applications such as autonomous driving and topographic surveying, offering enhanced reconstruction quality and improved processing efficiency. However, its ability to perform long-distance depth imaging raises potential privacy concerns, particularly in scenarios where individuals may be unknowingly captured. Addressing these concerns responsibly is essential for real-world deployment.